%% file: main.tex
\documentclass[conference]{IEEEtran}
\IEEEoverridecommandlockouts
\usepackage{cite}
\usepackage{amsmath,amssymb,amsfonts}
\usepackage{algorithmic}
\usepackage{graphicx}
\usepackage{textcomp}
\usepackage{xcolor}
\def\BibTeX{{\rm B\kern-.05em{\sc i\kern-.025em b}\kern-.08em
    T\kern-.1667em\lower.7ex\hbox{E}\kern-.125emX}}

\usepackage{balance}
\usepackage[utf8]{inputenc}
\usepackage{todonotes}
\usepackage{comment}
\usepackage{caption}
\usepackage{subcaption}
\usepackage{hyperref}
\usepackage{url}
\usepackage{amsmath}
\usepackage{enumitem}
\usepackage[english, ruled, onelanguage, vlined]{algorithm2e} 
\usepackage{etoolbox}
\usepackage{numprint}
\usepackage{diagbox}
\usepackage{MnSymbol,utfsym}
\usepackage{amsfonts}
\usepackage{multirow, array}
\usepackage{float}
\usepackage[dvipsnames]{xcolor}

\usepackage{tikz, pgfplots}
\usetikzlibrary{positioning, arrows.meta, automata, shapes.symbols,calc, graphs}
\makeatletter
\tikzoption{base font}{\def\tikz@base@textfont{#1}}
\tikzoption{font}{\def\tikz@textfont{\tikz@base@textfont#1}}
\tikzset{
  base font=\sffamily,
}

\newcommand\med{\text{med}}

\AtBeginDocument{%
  \providecommand\BibTeX{{%
    Bib\TeX}}}

\begin{document}

\title{Trivial~Graph~Features~and~Classical~Learning are~Enough~to~Detect~Random~Anomalies}

\author{
\IEEEauthorblockN{Matthieu Latapy}
\IEEEauthorblockA{\textit{CNRS, LIP6} \\
\textit{Sorbonne Université}\\
Paris, France \\
matthieu.latapy@lip6.fr}
\and
\IEEEauthorblockN{Stephany Rajeh}
\IEEEauthorblockA{\textit{Efrei Research Lab} \\
\textit{Efrei Paris Panthéon-Assas Université}\\
Villejuif, France \\
stephany.rajeh@efrei.fr}
}

\maketitle

\begin{abstract}
Detecting anomalies in link streams that represent various kinds of interactions is an important research topic with crucial applications. Because of the lack of ground truth data, proposed methods are mostly evaluated through their ability to detect randomly injected links. In contrast with most proposed methods, that rely on complex approaches raising computational and/or interpretability issues, we show here that trivial graph features and classical learning techniques are sufficient to detect such anomalies extremely well. This basic approach has very low computational costs and it leads to easily interpretable results. It also has many other desirable properties that we study through an extensive set of experiments. We conclude that detection methods should now target more complex kinds of anomalies.
\end{abstract}


\noindent
{\bf Reproducibility.}
We provide our code and data at \cite{TGFweb}.

\section{Introduction}
\label{sec:introduction}

Detecting anomalies is crucial in many domains such as finance, security, and social media. In many cases, detection relies on the analysis of interaction traces modeled as link streams \cite{latapy2018stream}: ordered sequences of timestamped links $(t_i,u_i,v_i)$ meaning that an interaction occurred between items $u_i$ and $v_i$ at time $t_i$. Items may represent bank accounts and links may represent payments, for instance; or they may represent digital devices and data transfers; or individuals and contacts between them.
Then, anomalous links typically reveal credit card frauds, forged packets or identity thefts.

Most detection methods need labeled data in which anomalies are known, either for training (supervised learning) or for evaluation (does the method succeed in detecting known anomalies?), or both. However, labeled data are rarely available. Then, researchers commonly rely on {\em anomalous link injection} \cite{akoglu2015graph}:
they add random links into real-world data and detection methods are designed for these links \cite{ranshous2016scalable,yoon2019fast, cai2021structural, bansal2022density, yu2018netwalk, zheng2019addgraph, chang2021f, sun2022monlad, liu2021anomaly, guo2023rustgraph}. 

We propose here a graph-based approach that favors simplicity. For each link, we consider history graphs composed of the most recent interactions. We then compute {\em trivial} features of each link within its history graphs, and we use them to train classical machine learning methods on real-world data with randomly injected anomalies. We call this method TGF, to emphasize the fact that it needs trivial graph features only.

Experiments show that this simple method detects injected links extremely well, better than the profusion of more complex state-of-the-art methods. In addition, it has the following important strengths, summarized in Table~\ref{TableStrengthsofOurApproach}.

{\em High flexibility.} All methods have some flexibility, but TGF has a clear advantage to this regard. In addition to a very flexible history management (details below), it is able to deal with various kinds of graphs (classical, bipartite, directed, etc) and it may use additional application-dependent features (a payment amount or location, for instance).



{\em Efficiency.}
TGF is very fast, both theoretically with algorithmic complexity in $\mathcal{O}(1)$ for all feature computations, and in practice with up to millions of links processed per second on a standard laptop. TGF space costs are also in $\mathcal{O}(1)$ and very small in practice. Importantly, space requirements are under control, as they may be tuned using history size and type to ensure they stay under any user-defined bound.

{\em Advanced history.} TGF uses past data at various resolutions, and it actually combines several resolutions very easily. It also uses several kinds of history graphs, and combines them. Other approaches for building history graphs may easily be added and combined. For instance one may use the graph of interactions that occurred during the same day one week ago. This ability to integrate different notions of history and time resolutions is quite unique.

{\em Interpretability.} Last but not least, trivial graph features have natural interpretations in terms of graphs. More importantly, they also have interpretation in terms of what they mean in the application area. As TGF readily identifies the features that play the most important role in anomaly detection, this leads to an interpretable description of what makes links anomalous in each practical case.

Having all these strengths combined in a simple method that detects randomly injected links so well leads us to the following conclusion: trivial graph features and classical machine learning are enough to detect anomalies, as long as they are similar to randomly injected links.

This is not a statement of important theoretical interest only: in many practical cases, it is reasonable to model anomalies by random links. Indeed, it is often impossible for fraudsters to mimic the real behavior of the legitimate user they pretend to be \cite{benson2018,wilsem2011,anderson2014,KHAN2021103112}.
In addition, most research work in anomaly detection is evaluated on random link injections, and therefore much effort is devoted to the optimization of detection in such cases. We believe that TGF puts an end to this line of research.

\section{Related work} \label{sec:related-work}

Link streams, also called edge streams, temporal graphs, temporal networks, or dynamic graphs by various authors, model a wide variety of interaction traces. Detecting anomalous links in such streams is an important fundamental and applied research topic.

Many recent approaches leverage deep learning methods \cite{grover2016node2vec, perozzi2014deepwalk, yu2018netwalk, zheng2019addgraph, cai2021structural, zong2018deep, chen2020anomaly}. For instance, Node2Vec \cite{grover2016node2vec}, DeepWalk \cite{perozzi2014deepwalk}, and others use embeddings based on random walks (with the goal that similar neighborhoods lead to similar embeddings) and then apply classical clustering methods. NetWalk \cite{yu2018netwalk} learns dynamic latent node representations and uses auto-encoder reconstruction error. AddGraph \cite{zheng2019addgraph}, StrGNN \cite{cai2021structural} and others use graph convolutional networks (GCN) and graph neural network to extract local structural and temporal features. TADDY \cite{liu2021anomaly} proposes a transformer-based method, while RustGraph \cite{guo2023rustgraph} combines variational autoencoding and contrastive learning. Using two self-supervised tasks, SLADE \cite{lee2024slade} learns dynamic node representations by updating memory vectors for short-term stability and reconstructing them from recent interactions. DAGMM \cite{zong2018deep} provides an unsupervised solution that uses a deep autoencoder and a Gaussian mixture model. GraphSAGE \cite{hamilton2017inductive} leverages node embeddings to perform semi-supervised representation learning. BEA \cite{chen2020anomaly} deals with bipartite networks; it detects burstiness and models dynamic connectivity patterns using embeddings.
All these methods reach very good detection scores, but they often raise serious interpretability and scalability issues.

Direct graph-based approaches also exist \cite{akoglu2010oddball, eswaran2018sedanspot, ranshous2016scalable, bhatia2020midas}. For instance, SedanSpot \cite{eswaran2018sedanspot} samples the whole graph and uses random walks to assess the impact of new links on the distance between sampled nodes. CM-Sketch \cite{ranshous2016scalable} is a sketch-based method that leverages local structural information and historical behavior in the vicinity of links to classify them as anomalous or not. These methods often provide better interpretability, but they generally use strong assumptions on anomaly features that weaken their flexibility. 


Because of the lack of publicly available data with known anomalies that would provide ground truth, evaluation of proposed methods generally relies on injected anomalies \cite{akoglu2015graph}. Then, many detection methods focus on the detection of randomly injected links \cite{ranshous2016scalable,yoon2019fast, cai2021structural, bansal2022density, yu2018netwalk, zheng2019addgraph, chang2021f, sun2022monlad, liu2021anomaly, guo2023rustgraph}.

\begin{table}[h]
\centering
\caption{Comparison of TGF with state-of-the-art methods.}
\resizebox{.98\columnwidth}{!}{\begin{tabular}{|c|cccc|}
\hline
\backslashbox{Method}{Strength}
&\makebox[5.7em]{\rotatebox[origin=c]{30}{High flexibility}}
&\makebox[5.7em]{\rotatebox[origin=c]{30}{Efficiency}}
&\makebox[5.7em]{\rotatebox[origin=c]{30}{Advanced history}}
&\makebox[5.7em]{\rotatebox[origin=c]{30}{Interpretability}}\\
\hline
SedanSpot & & \usym{1F5F8} &  &  \usym{1F5F8} \\\hline
CM-Sketch &   & \usym{1F5F8} &  &  \usym{1F5F8} \\\hline
Node2Vec & \usym{1F5F8} &   &  &  \\\hline
DeepWalk & \usym{1F5F8} &  &  &   
\\\hline
NetWalk & \usym{1F5F8} & \usym{1F5F8} & \usym{1F5F8} &  
\\\hline
AddGraph & \usym{1F5F8} &   & \usym{1F5F8} &   
\\\hline
StrGNN &   &    & \usym{1F5F8} &  \\\hline
TADDY &  &   &  \usym{1F5F8}  &   
\\\hline
RustGraph &  &   & \usym{1F5F8}  &    
\\\hline
SLADE &  &   & \usym{1F5F8} &    
\\\hline
DAGMM & \usym{1F5F8} &  &  & 
\\\hline
GraphSAGE & \usym{1F5F8}  & \usym{1F5F8} &  & 
\\\hline
BEA &  &  &  \usym{1F5F8} &  
\\\hline
\hline
\textbf{TGF} & \usym{1F5F8} & \usym{1F5F8}  &  \usym{1F5F8}  & \usym{1F5F8}
\\\hline
\end{tabular}}
\label{TableStrengthsofOurApproach}
\end{table}

In contrast with above methods, we propose a lightweight and easily interpretable method. Despite its striking simplicity, it outperforms all above methods for detecting randomly injected links. In addition, it has a unique set of strengths that makes it particularly appealing, see Table~\ref{TableStrengthsofOurApproach}.

Notice that some works use datasets (in general only one) with labeled anomalies. These datasets however are private in most cases, which makes comparison impossible.
Some works target specific application domains, like network traffic or financial transactions, and they leverage domain-specific features, like protocol types or payment amounts. Instead, we deal with $(t,u,v)$ triplets only. 
Finally, some methods target complex anomalies like clusters of links, while we target individual anomalous links.

We do not detail these works because they are out of the scope of this paper: our goal is to show that, although randomly injected links are widely used in the literature, they actually are easy to detect.

\section{History graphs}
\label{sec:history-graphs}

A link is not normal or anomalous in itself; it is so {\em in a given context}. This context consists of past links, which reflect for instance usual credit card use, typical packet transfers, or ordinary user behaviors. In addition, recent links are generally considered more significant than older ones.

\begin{figure}[h]
\centering
\resizebox{.9\columnwidth}{!}{\input{history.tikz}}
\caption{
\textbf{Examples of $G$-type and $H$-type history graphs.} Top: a link stream between 4 nodes $a$, $b$, $c$, and $d$, from time $0$ to $10$. We consider the latest link $(10,b,c)$, meaning that an interaction occurred between $b$ and $c$ at time $10$. We display two $G$-type ($G_3$ and $G_8$, bottom-left) and two $H$-type ($H_3$ and $H_8$, bottom-right) history graphs for this link. The integers on the links of these graphs indicate their number of occurrences within the considered history. For instance, $H_3$ is the graph obtained from the $3$ last interactions. They involve $c$ and $d$ twice and $a$ and $b$ once.
}
\label{fig:history}
\end{figure}
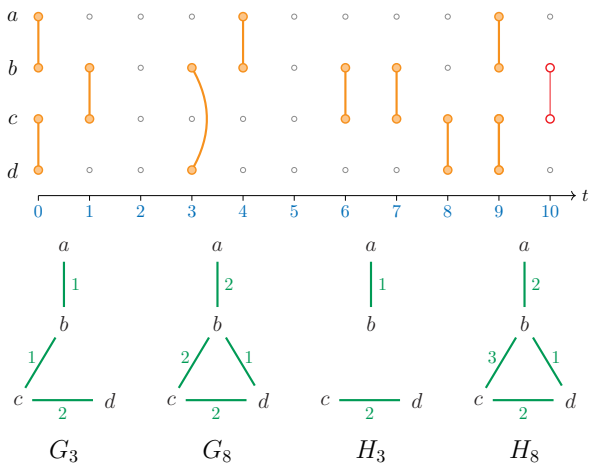

It is therefore common to analyze link properties within the graph $G_d$ of links occurring during the last time period of duration $d$, for a given $d$. See Figure~\ref{fig:history}. However, selecting an appropriate duration is challenging and it can profoundly influence results. Additionally, adopting a single fixed duration may lead to missing critical information embedded in the network dynamics, which generally spans several time scales.

In this paper, we consider a collection $\Delta$ of durations instead of just one. We then obtain for each link a set of several history graphs representing past activity, one for each value $d$ in $\Delta$. We compute the features of each link with respect to each of these history graphs. By taking different time scales in $\Delta$, from small ones to much larger ones, these history graphs capture rich stream dynamics. 

With this approach, we obtain history graphs of several prescribed durations. However, the size of these graphs in terms of numbers of nodes and links is unbounded: they may contain one link per interaction in the considered period. Then, activity bursts lead to huge history graphs, which raises two issues: memory needed for storing them may sometimes be prohibitive; and we compute features within graphs of different sizes, making them difficult to compare.

In order to tackle such situations, we also consider a collection $\Sigma$ of history sizes $s$. For each size $s$ in $\Sigma$ we consider the history graph $H_s$ of the last $s$ links occurring before it, see Figure~\ref{fig:history}. We then compute features of each new link within these history graphs. In this way, the size of the considered graphs stays under control (they contain at most $s$ links), thus needed memory and computation time are bounded.

More formally, we consider a link sequence $S = (t_1,u_1,v_1),\allowbreak\ (t_2,u_2,v_2),\ \dots$ in which $t_i\le t_{i+1}$ for all $i$, and for any link $(t_i,u_i,v_i)$ in $S$, we introduce two kinds of history graphs:
\begin{itemize}[noitemsep,topsep=0pt,parsep=0pt,partopsep=0pt,leftmargin=.5\leftmargin]
\item
the $d$-duration $G$-type history graph $G_d=(V_d,E_d)$, defined for a given duration $d$ by $E_d = \{(u,v), \exists (t_j,u,v) \in S, j<i, t_i-t_j\le d\}$ and $V_d$ the set of nodes involved in $E_d$; 
\item
the $s$-size $H$-type history graph $H_s=(V_s,E_s)$, defined for a given size $s$ by $E_s = \{(u,v), \exists (t_j,u,v) \in S, j<i, i-j\le s\}$ and $V_s$ the set of nodes involved in $E_s$.
\end{itemize}

\begin{algorithm}[h]
\caption{Browsing $G$-type history graphs}
\label{algo:browse-history}
\KwIn{sequence $S$ of time-ordered links, duration $d$}
\KwOut{the features of $S$ links in $G_d$}
create an empty queue $Q$\;
\ForAll{$(t,u,v)$ in $S$}{
 add $(t,u,v)$ to $Q$\;
 create/increase counter named $(u,v)$\;
 \While{the first element $(t',u',v')$ in $Q$ verifies $t-t'>d$}{
  decrease/delete counter named $(u',v')$ \;
  remove $(t',u',v')$ from $Q$\;
  }
 output features of $(u,v)$ in $G_d$\;
 }
\end{algorithm}

Algorithm~\ref{algo:browse-history} shows how to browse $G$-type history graphs. The algorithm for $H$-type history graphs is very similar and we do not detail it. 
In this pseudo-code, {\em create/increase} means that we create a counter with value $0$ if it is not already existing, and we increment it in all cases. Conversely, {\em decrease/delete} means that we decrease the counter and delete it if it reaches the $0$ value. 

This algorithm only needs a counter named after each link in the current graph; no additional data structure is needed. This set of counters is handled in $\mathcal{O}(1)$ average time and space thanks to a hash table. As a consequence, handling each link in the stream has a cost independent of the chosen history size.


Interestingly, this algorithm does not only browse the wanted graphs; it also stores the number of occurrences of each link, see Figure~\ref{fig:history}. This number is required for the correct update of history graphs, since a link is removed only when its number of occurrences reaches zero. As a side effect, we have {\em weighted} history graphs that capture richer information on the context of each link. We will intensively use these weights in the following.

\section{Trivial graph features}
\label{sec:graph-features}

This section presents a wide set of trivial features to characterize a given link in a given graph. These features are {\em trivial} in the following sense:
\begin{itemize}[noitemsep,topsep=0pt,parsep=0pt,partopsep=0pt,leftmargin=.5\leftmargin]
\item they are easy to interpret because they capture natural and intuitive properties of interactions, like their frequency,
\item they only deal with local information in the graph, at the level of individual nodes and links,
\item their computational cost is in $\mathcal{O}(1)$ time and space, making them particularly suitable for large-scale networks.
\end{itemize}

However, {\em computing} these trivial features with such a small time and space cost is not trivial. It is possible only thanks to a data structure that we call {\em decreasing sorted counters} (DSC). It maintains a sorted array of counters, with counter creation, deletion, increase and decrease in $\mathcal{O}(1)$ time and space. Then, it gives the number of counters having a given value, the maximum counter value, and the median counter value in $\mathcal{O}(1)$ time and space.
Since DSC is considered folklore \cite{stack}, and since similar principles are implicitly used in some previous papers \cite{DSCcores2011,DSCcores2003}, we only present DSC details and implementation in the online appendix \cite{TGFweb}.


\label{sec:O1unipartite}

Our goal here is to characterize an occurrence of a link $(u,v)$ in a history graph $G=(V,E)$ with $E\subseteq V\times V$. This history graph is an undirected weighted graph with weight function $\omega$ giving the number of occurrences of each link in the considered history. If a link is not in $E$, then we consider that its weight is $0$. Since $G$ is undirected, there is no distinction between $(u,v)$ and $(v,u)$.

For any $x \in V$, we denote by $N(x) = \{y, (x,y)\in E\}$ the neighborhood of $x$, by $d(x) = |N(x)|$ the degree of $x$, and by $\delta(x) = \sum_y \omega(x,y)$ its weighted degree.

\noindent
A first family of features describes the graph $G$:
\begin{itemize}[noitemsep,topsep=0pt,parsep=0pt,partopsep=0pt,leftmargin=.5\leftmargin]
\item
the number of nodes $n=|V|$, the number of links $m=|E|$, and the total weight $\mu = \sum_{x,y} \omega(x,y)$,
\item
the number of degree $1$ nodes $|\{x , d(x) = 1\}|$ in $G$, the number of degree $2$ nodes $|\{x , d(x) = 2\}|$, the maximal degree $\max_{x \in V}(d(x))$, and the median degree $\med_{x \in V}(d(x))$,
\item
the number of weighted degree $1$ nodes $|\{x , \delta(x) = 1\}|$, the number of weighted degree $2$ nodes $|\{x , \delta(x) = 2\}|$, the maximal weighted degree $\max_{x \in V}(\delta(x))$, and the median weighted degree $\med_{x \in V}(\delta(x))$,
\item
the number of links with weight $1$, the number of links with weight $2$, the maximal link weight, and the median link weight.
\end{itemize}

\noindent
A second family of features focuses on the properties of link $(u,v)$ within $G$ (since we consider undirected links, we assume without loss of generality that $d(u)\le d(v)$): 
\begin{itemize}[noitemsep,topsep=0pt,parsep=0pt,partopsep=0pt,leftmargin=.5\leftmargin]
\item the degrees $d(u)$ and $d(v)$ of $u$ and $v$,
\item the number of nodes having degree $d(u)$ and $d(v)$, and the number of nodes having a degree larger than $d(u)$ and $d(v)$,
\item the weighted degrees $\delta(u)$ and $\delta(v)$ of $u$ and $v$,
\item the number of nodes having weighted degree $\delta(u)$ and $\delta(v)$, and the number of nodes having a weighted degree larger than $\delta(u)$ and $\delta(v)$,
\item the weight $\omega(u,v)$ of $(u,v)$,
\item the number of links having weight $\omega(u,v)$ and the number of links having a weight larger than $\omega(u,v)$.
\end{itemize}

These features are very easy to interpret. For instance, if the considered data consist in contacts between individuals, then the degree of a node is the number of individuals met by this individual, its weighted degree is the number of contacts it had, and the weight of a link is the number of contacts between the two corresponding individuals.


Notice also that having features for both the graph and the considered link is important for learning and feature interpretation. For instance, even if the degree of a node is meaningful in itself, it is interesting to compare it to the average, median, and maximal degrees in the graph, and to compare it to the total number of nodes.

Thanks to the decreasing sorted counter (DSC) structure, we compute all features above very efficiently. We use two counters for each node: one for its degree and one for its weighted degree. We also use a counter for each link weight. These counters are managed by three DSC structures: one for node degrees, one for node weighted degrees, and one for link weights. Then, handling each link in the stream induces a finite number of counter increasing/creation or decreasing/deletion in DSC structures, all in $\mathcal{O}(1)$ time and space. Access to the wanted features from these structures also has a constant computational cost, see online appendix and code \cite{TGFweb} for implementation details.

\section{Experimental setup}
\label{sec:experimental-setup}

In order to evaluate TGF and compare it to state-of-the-art methods, we need three elements: real-world datasets of various types and scales, an anomaly injection method, and a learning technique to classify input links using the computed features. We detail these three aspects below in order to ensure our work is easily reproducible.

\subsection{Datasets}
 
We use first a family of widely used and publicly available reference datasets, described in \cite{kunegis2013konect}. They are named  Bitcoin-Alpha, Bitcoin-OTC, DNC Emails, UCI Messages, Digg, and Internet Topology (first six rows in Table~\ref{TableBenchmarkDatasets}).

These datasets already span well the variety of anomaly detection application areas. However, we are also interested in large-scale performances of TGF. We therefore add the following datasets (last three rows in Table~\ref{TableBenchmarkDatasets}).

\begin{table}[h]
\centering
\caption{
Basic characteristics of datasets used in this paper, ordered with respect to their total number of timestamped links $\ell$ (last column). The other columns indicate
the number of nodes $n$, and the number of distinct links $m$.
}
\resizebox{.9\columnwidth}{!}{\begin{tabular}{l|r|r|r}
Dataset &  $n$ & $m$ & $\ell$ \\
\hline
Bitcoin-Alpha &  3,783  & 14,124   & 24,186   \\
Bitcoin-OTC & 5,881  & 21,492   & 35,592   \\ 
DNC Emails & 1,866  & 5,517 & 37,421    \\
UCI Messages &  1,899 & 13,838 & 59,835   \\
Digg & 30,360  & 85,151  & 86,199   \\
Internet Topo. & 34,761  & 107,720   & 171,403    \\
Taxi & 320,413 & 52,254,156 & 153,268,910 \\
Mawi & 2,170,552 & 6,916,241 & 223,234,909 \\
Bitcoin-BC & 699,834,203 & 1,334,954,478 & 1,515,990,044
\end{tabular}}
\label{TableBenchmarkDatasets}
\end{table}

{\bf Mawi} is based on a publicly available 24-hours long internet packet-level traffic trace collected in 2019\footnote{http://mawi.wide.ad.jp/mawi/ditl/ditl2019-G/} \cite{mawilab}. We keep UDP and TCP traffic only and we round each timestamp to a one second precision. We then represent each packet transfer by a link between its source and destination IP addresses.

{\bf Taxi} is built from a publicly available recording of 165,114,362 taxi trips in New-York city in 2014\footnote{\url{https://data.cityofnewyork.us/Transportation/2014-Yellow-Taxi-Trip-Data/gkne-dk5s}}. We split the NYC region into a grid of $1000\times 1000$ cells that we call nodes, and consider that each trip starting in a cell and ending in another cell makes a link between corresponding nodes. We timestamp each link with the average between its trip starting and ending times, in seconds.

{\bf Bitcoin-BC} is built from the publicly provided set of 623,483,734 transactions recorded in the Bitcoin blockchain from its beginning on January 3, 2009 until March 13, 2021 \cite{jules}. In order to keep the most significant transactions only, we ignore the ones with 10 or more input or output addresses. Then, we build the link stream in which $(t,u,v)$ means that $u$ occurred as an input address of a transaction in block of index $t$ and $v$ was an output address of this transaction.

\subsection{Anomaly injection}
\label{sec:injection}

Since our approach relies on supervised learning, we need labeled data in which each link is tagged as normal or anomalous. However, most available datasets do no provide such information. Then, the classical approach consists in injecting random links considered as anomalous, original links being considered as normal \cite{akoglu2015graph}.

More formally, injected links are generated as follows. We consider a link sequence $S = (t_1,u_1,v_1),\allowbreak\ (t_2,u_2,v_2),\ \dots,\ (t_\ell,u_\ell,v_\ell)$. We denote by $T = \{t_i,\ i=1, 2, \dots, \ell\}$ the set of all timestamps, and by $V = \{u_i,\ i=1, 2, \dots, \ell\} \cup \{v_i,\ i=1, 2, \dots, \ell\}$ the set of all nodes.
Then, each injected link is built by sampling uniformly at random a timestamp in $T$ and two distinct nodes in $V$, such that $(t,u,v)$ does not appear in $S$. 

We follow the classical convention that consists in injecting a number of anomalous links equal to 1\%, 5\%, and 10\% of the total number of original links $\ell$.

\subsection{Learning method} \label{sec:learning}

In our context, the learning deals with a sequence of timestamped links $S = s_1, s_2, \dots, s_\ell$ of $\ell$ triplets $s_i = (t_i,u_i,v_i)$, with $t_i\le t_{i+1}$ for all $i$. Each link is labeled as normal or anomalous, and TGF characterizes each link with a list of features, as explained above. Notice that one may add domain-specific features to this list, depending on the application area, and apply the following with no change.

From the list of features associated to each link, one may use any supervised learning method to produce a classifier to detect anomalies. This gives a high flexibility to our approach.
In this paper, we mostly use the scikit-learn implementation of the random forest classifier \cite{breiman2001random} with default settings\,\footnote{These settings are detailed at \url{https://scikit-learn.org/stable/modules/generated/sklearn.ensemble.RandomForestClassifier.html}}. This learning method is known to effectively deal with large numbers of features, to be robust to noise in the data, and to resist over-fitting.
We will compare it to other learning methods (namely, gradient boosting classifier and SVM) in Section~\ref{sec:inpractice}. We will also discuss fine tuning of learning parameters in that section.

Learning relies on two sub-sequences extracted from $S$: a training set $X$ and a test set $Y$.
We follow here a classical method, which is the most widely used in the literature. Let us consider a ratio $r$ in $[0,1]$ and let us denote by $k$ the value $r\cdot\ell$. This classical method defines $X$ and $Y$ as the sub-sequences $s_1,s_2,\dots, s_k$ and $s_{k+1},\dots,s_\ell$ respectively. For instance, $r = 0.8$ means the training set is composed of the $80 \%$ first links, and the test set of the $20 \%$ remaining links. In our experiments, we will use $r$ in $\{0.7, 0.8, 0.9\}$.

Finally, for any given learning set $X$ and test set $Y$, we proceed as follows. We deal with class imbalance (the vast majority of links are normal) using undersampling of $X$: we select all anomalous links in $X$ and the same number of randomly chosen normal links in $X$. This is known to improve learning quality and to make learning more scalable \cite{sun2009classification}. We apply the learning method to this subset of $X$ and obtain a classifier.

Then, we run this classifier on the test set $Y$, and we use ROC-AUC to assess performances, because it is the prevalent criterion in the literature. It is the area under the receiver operating characteristic curve, which plots the true positive rate against the false positive rate at various classification thresholds. A higher AUC score indicates better discrimination between positive and negative classes.

We also performed $k$-fold cross validation learning (with $k$ = 10) in order to gain more insight. However, results are very similar to the ones of the train/test split above, therefore we do not detail them.

\section{Experimental results}
\label{sec:experimental-results}

In this section, we first compare TGF to state-of-the-art methods. We then explore several key aspects of TGF:
its ability to take benefit from several history graphs, how to typically use it in practice, its interpretability capabilities, and finally its computational costs. 

\subsection{Comparison to state-of-the-art}

In order to assess TGF performances, we compare it to a wide variety of state-of-the-art methods applied to public reference datasets. We reproduce in Table~\ref{TableComparisonWithStateOfTheArt-Unipartite} the main previously published results \cite{liu2021anomaly,cai2021structural, chen2020anomaly, guo2023rustgraph} and add TGF results in two configurations: when it uses only the $H$-type history graph of size $1000$, and when it uses a combination of several history graphs, which we detail in next section. We use the classical learning technique detailed in Section~\ref{sec:learning} with parameter $r=0.7$, meaning that we use the $70\%$ first links for training and the $30\%$ remaining ones for testing. These results are typical of results obtained with wide ranges of parameter values, that we will explore in depth in subsequent sections.

\begin{table*}[h]
\centering
\caption{AUC performances of state-of-the-art methods and TGF in two configurations: with $H$-type history graphs of size $1000$, and with combined history graphs. All methods are applied to public reference datasets with $1$\%, $5$\% and $10$\% anomaly injection. The two best results are in bold, the best one is underlined.}
\resizebox{\textwidth}{!}{\begin{tabular}
{@{}l|c@{\hspace{0.05in}}c@{\hspace{0.05in}}c|c@{\hspace{0.05in}}c@{\hspace{0.05in}}c|c@{\hspace{0.05in}}c@{\hspace{0.05in}}c|c@{\hspace{0.05in}}c@{\hspace{0.05in}}c|c@{\hspace{0.05in}}c@{\hspace{0.05in}}c|c@{\hspace{0.05in}}c@{\hspace{0.05in}}c@{}} 
& \multicolumn{3}{c|}{UCI Messages} 
& \multicolumn{3}{c|}{Digg}
& \multicolumn{3}{c|}{DNC Emails}
& \multicolumn{3}{c|}{Bitcoin-Alpha}
& \multicolumn{3}{c|}{Bitcoin-OTC}
& \multicolumn{3}{c}{Internet Topology}
\\
& 1\% & 5\% & 10\% & 1\% & 5\% & 10\%
& 1\% & 5\% & 10\% & 1\% & 5\% & 10\%
& 1\% & 5\% & 10\% & 1\% & 5\% & 10\%
\\
SedanSpot
& 0.7342 & 0.7156 & 0.7061 & 0.6976 & 0.6784 & 0.6396
& 0.7427 & 0.7362 & 0.7235 & 0.7380 & 0.7264 & 0.7085
& 0.7346 & 0.7284 & 0.7156 & 0.6873 & 0.6742 & 0.6672
\\ 
CM-Sketch
& 0.7320 & 0.6968 & 0.6835 & 0.6884 & 0.6675 & 0.6358
& 0.7053 & 0.6946 & 0.6876 & 0.7146 & 0.7015 & 0.6887
& 0.7412 & 0.7338 & 0.7242 & 0.6687 & 0.6605 & 0.6558
\\ 
Node2Vec
& 0.7371 & 0.7433 & 0.6960 & 0.7364 & 0.7081 & 0.6508
& 0.7391 & 0.7284 & 0.7103 & 0.6910 & 0.6802 & 0.6785
& 0.6951 & 0.6883 & 0.6745 & 0.6821 & 0.6752 & 0.6668
\\  
DeepWalk
& 0.7514 & 0.7391 & 0.6979 & 0.7080 & 0.6881 & 0.6396
& 0.7481 & 0.7303 & 0.7197 & 0.6985 & 0.6874 & 0.6793
& 0.7423 & 0.7356 & 0.7287 & 0.6844 & 0.6793 & 0.6682
\\ 
NetWalk
& 0.7758 & 0.7647 & 0.7226 & 0.7563 & 0.7176 & 0.6837
& 0.8105 & 0.8371 & 0.8305 & 0.8385 & 0.8357 & 0.8350
& 0.7785 & 0.7694 & 0.7534 & 0.8018 & 0.8066 & 0.8058
\\
AddGraph
& 0.8083 & 0.8090 & 0.7688 & 0.8341 & 0.8470 & 0.8369
& 0.8393 & 0.8627 & 0.8773 & 0.8665 & 0.8403 & 0.8498
& 0.8352 & 0.8455 & 0.8592 & 0.8080 & 0.8004 & 0.7926
\\
StrGNN
& 0.8179 & 0.8252 & 0.7959 & 0.8162 & 0.8254 & 0.8272
& 0.8775 & 0.9103 & 0.9080 & 0.8574 & 0.8667 & 0.8627
& 0.9012 & 0.8775 & 0.8836 & 0.8553 & 0.8352 & 0.8271
\\
TADDY
& 0.8912 & 0.8398 & 0.8370 & 0.8617 & 0.8545 & 0.8440
& 0.9348 & 0.9257 & 0.9210 & 0.9451 & 0.9341 & 0.9423
& 0.9455 & 0.9340 & 0.9425 & 0.8953 & 0.8952 & 0.8934
\\
RustGraph
& 0.9128 & 0.9117 & 0.9124 & \textbf{0.8795} & \textbf{0.8577}  & \textbf{0.8624}
& 0.9855 & 0.9837 & 0.9806 & 0.9477 & 0.9348 & 0.9207
& 0.9608 & 0.9578  & 0.9603 & \textbf{0.9746} & \textbf{0.9768} & \underline{\textbf{0.9761}}
\\ 
SLADE
& 0.7847 & 0.7894 & 0.7715 & 0.8026 & 0.7889 & 0.7865
& 0.9159 & 0.9065 & 0.9109 & 0.9009 & 0.9137 & 0.9188
& 0.9262 & 0.9306 & 0.9198 & 0.9282 & 0.9295 &  0.9281
\\
\textbf{TGF $H_{1000}$}
& \textbf{0.9580} & \textbf{0.9635} & \textbf{0.9646} & 0.8408 & 0.8418  & 0.8382
& \textbf{0.9892} & \textbf{0.9893} & \textbf{0.9891} & \textbf{0.9662} & \textbf{0.9633} & \textbf{0.9624}
& \textbf{0.9686} & \textbf{0.9739}  & \textbf{0.9732} & 0.9639 & 0.9637 & 0.9568
\\ 
\textbf{TGF comb.}
& \underline{\textbf{0.9857}} & \underline{\textbf{0.9876}} & \underline{\textbf{0.9872}}
& \underline{\textbf{0.9071}} & \underline{\textbf{0.8988}} & \underline{\textbf{0.8921}}
& \underline{\textbf{0.9919}} & \underline{\textbf{0.9952}} & \underline{\textbf{0.9958}} 
& \underline{\textbf{0.9820}} & \underline{\textbf{0.9942}} & \underline{\textbf{0.9923}}
& \underline{\textbf{0.9908}} & \underline{\textbf{0.9867}} & \underline{\textbf{0.9863}} 
& \underline{\textbf{0.9823}} & \underline{\textbf{0.9807}} & \underline{\textbf{0.9761}}
\\ 
\end{tabular}}
\label{TableComparisonWithStateOfTheArt-Unipartite}
\end{table*}

TGF scores are outstanding: in almost all cases, TGF is over $0.95$ in its $H_{1000}$ version, and over $0.98$ in its combined version. It is over $0.99$ in one third of all cases. Digg is a notable exception, but even in this case TGF scores are above $0.89$.
We insist on the fact that these results are obtained without any parameter tuning: we use the same default learning settings for all datasets.

With these scores, TGF consistently outperforms almost all state-of-the-art methods, for all injection rates, and on all datasets, even with only one history graph of fixed size. RustGraph however is a serious challenger: it outperforms TGF with $H_{1000}$ for the Digg and Internet Topology datasets. But TGF takes the lead with combined history graphs in all cases (scores are equal for Internet Topology with 10\% injection rate, rightmost column in Table~\ref{TableComparisonWithStateOfTheArt-Unipartite}).

More importantly, except in the Digg case, obtained scores are very close to the maximal possible value, $1$. This seriously questions the relevance of trying to improve detection scores even more.
Instead, we insist on the fact that performance improvement is not the unique strength of TGF: most importantly, TGF is very simple and efficient, whereas the best competing methods often rely on complex and costly approaches that are hard to interpret.

\subsection{Using diverse resolutions}
\label{sec:combination}

TGF offers several ways to use contextual information in the form of $H$-type or $G$-type history graph features at various resolutions. In addition, TGF is able to combine them, thus making it possible to use information at various time scales, and even to combine $H$- and $G$-type features. This is an important strength that allow for flexible consideration of both recent and distant past contexts.

\begin{figure}[h]
\centering
\resizebox{\columnwidth}{!}{\includegraphics{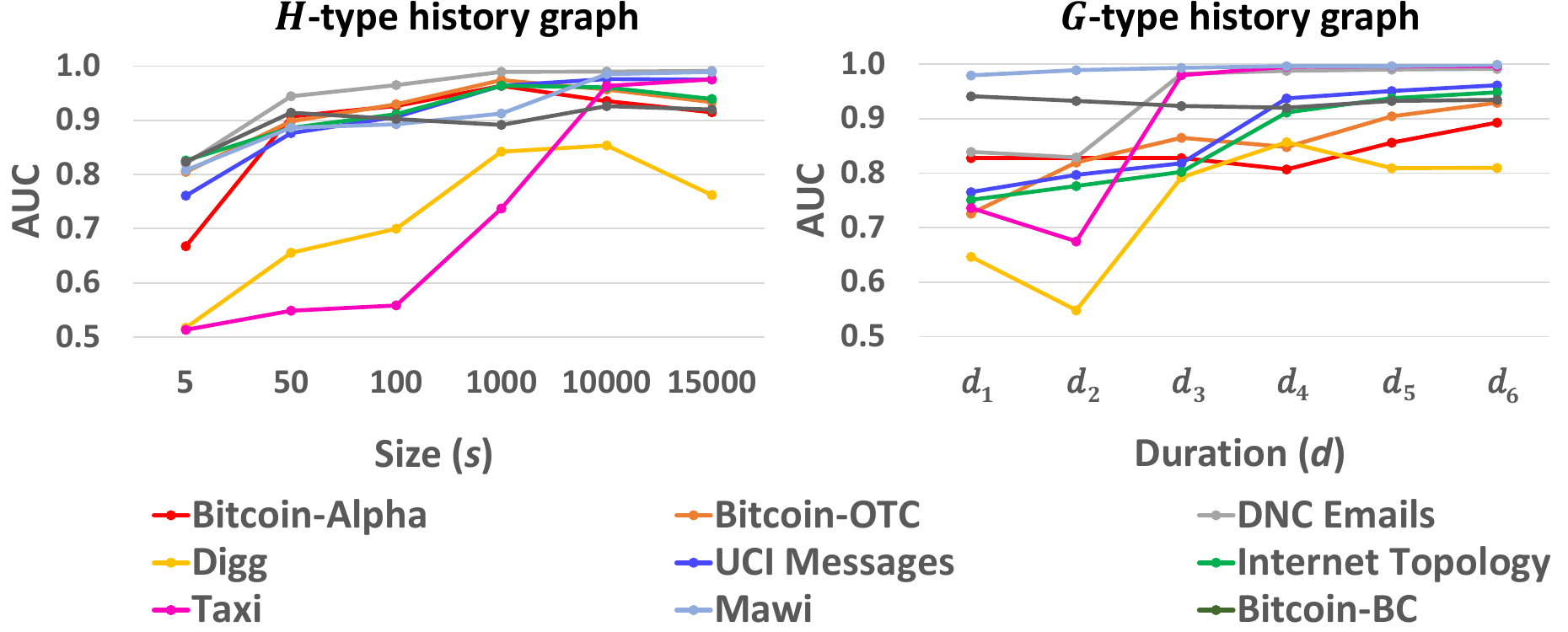}}
\caption{The impact of size $s$ and duration $d$ (horizontal axis) of the $H$-type (left) and $G$-type (right) history graphs on AUC scores. We consider here $5$\% anomaly injection and learning with $r=0.7$.}
\label{figAggregation}
\end{figure}

Figure~\ref{figAggregation} displays the impact of history size and duration on results for all our datasets, including the large-scale ones (Taxi, Mawi, and Bitcoin-BC), each with $5$\% anomaly injection and learning with $r=0.7$.
Other injection rates and parameters yield similar results, so they are not detailed.

In this experiment, we assess the performance of TGF by setting various $s$ sizes and $d$ durations for history graphs $H_s$ and $G_d$, respectively. More precisely, $s$ is set to the values $5$, $50$, $100$, $1000$, $10000$, and $15000$ links; while $d$ is set to $d_1=5$ seconds, $d_2=1$ minute, $d_3=1$ hour, $d_4=1$ day, $d_5=3$ days, and $d_6=6$ days.

There are two exceptions, though. Since Bitcoin-BC timestamps are expressed in blocks that may contain many of links, we use $d_1=1$, $d_2=3$, $d_3=6$, $d_4=12$, $d_5=24$, and $d_6=72$ blocks. As a block is added approximately every $10$ minutes, this leads to
$d_1\sim10$ minutes, $d_2\sim0.5$ hour, $d_3\sim1$ hour, $d_4\sim2$ hours, $d_5\sim4$ hours, and $d_6\sim 12$ hours.
The case of Mawi is different: it lasts 24 hours only, with typically thousands of links per second. Therefore, we set $d_1=1$ second, $d_2=5$ seconds, $d_3=10$ seconds, $d_4=0.5$ minute, $d_5=1$ minute, and $d_6=5$ minutes.

Figure~\ref{figAggregation} clearly shows that TGF rapidly reaches excellent AUC scores, often better than the ones presented in previous section with $H_{1000}$, as soon as history size or duration is not too small. For instance, we obtain a $0.73$ AUC score for Taxi with $H_{1000}$ and a $0.97$ AUC score with $H_{15000}$. Notice however that the scores are not necessarily better with larger or longer history graphs, but the difference is small in general: the worst case by far is Digg, which has a $0.84$ AUC score with $H_{1000}$ and a $0.76$ AUC score with $H_{10000}$. Results with $G$-type history graphs show similar behaviors.

\begin{figure}[h]
\centering
\resizebox{\columnwidth}{!}{\includegraphics{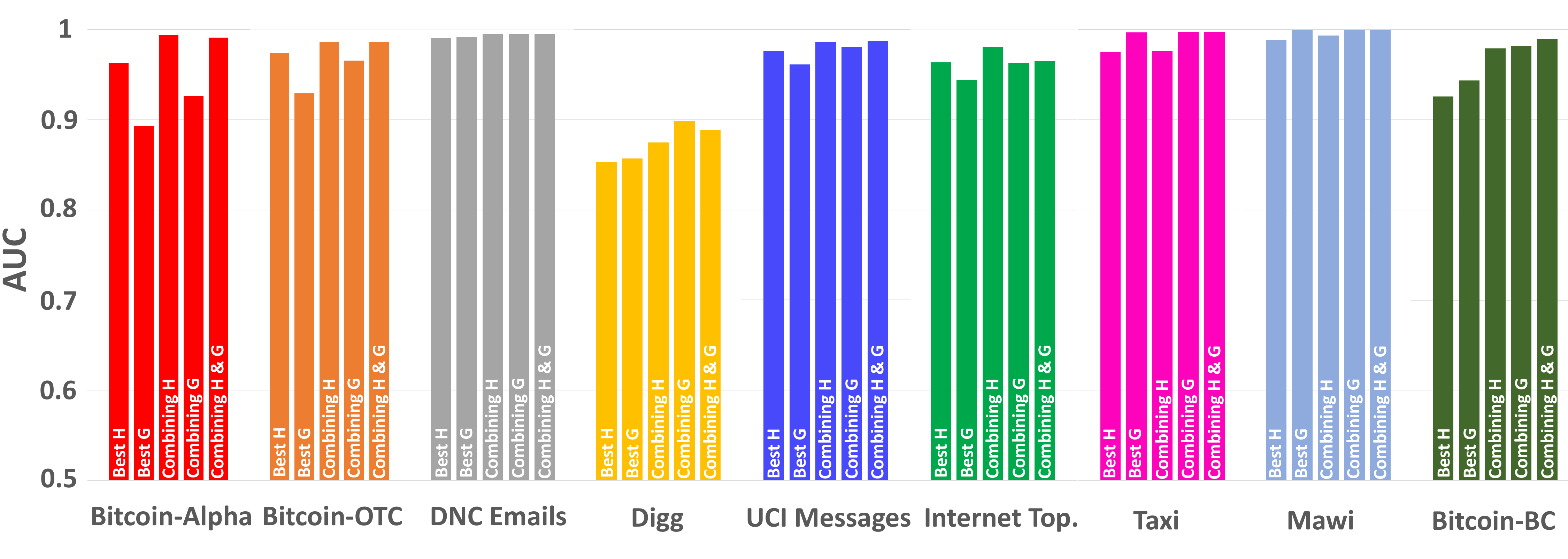}}
\caption{AUC scores obtained with various history resolutions and combinations. For each dataset, we display the best score obtained with, from left to right: a $H$-type history graph, a $G$-type history graph, the combination of all $H$-type history graphs, the combination of all $G$-type history graphs, and the combination of all these history graphs.
We consider 5\% anomaly injection and learning with $r=0.7$.}
\label{figCombining}
\end{figure}

Figure~\ref{figCombining} goes further by considering combinations of history graphs. The best results are listed in Table~\ref{TableBenchmarkDatasets} (last row). Whereas experiments above used only one type of history graph of only one size or duration, we consider here five cases: the best AUC score obtained with one of the $H$-type or one of the $G$-type history graphs above, the best score obtained with the combination of all the $H$-type or the $G$-type history graphs above, and the best score obtained with the combination of all these graphs.

These experiments consistently show that combining several history graphs of the same type improves results, sometimes very significantly. In some cases, $H$-type history graphs lead to better performances than $G$-type ones, but in other cases they are outperformed by $G$-type history graphs. This shows that considering several kinds of history graphs is important, with non-trivial effects. Finally, combining all graphs often gives the best results, but this is not true for all datasets.


\subsection{TGF with sliding windows}

Results presented until now work on whole datasets, divided into large training and test sets. This raises two concerns. First, since the considered data have a temporal nature, it may be more relevant to perform learning on recent data only. Second, this training uses huge amounts of data, which may have prohibitive computational costs. In this section, we explore the ability of TGF to perform detection using smaller and most recent data only.

To do so, for each dataset $S$, we first consider sliding windows that contain half the links of $S$. For a given $\Delta$ and for any $i = 0..\left\lfloor \frac{\ell/2}{\Delta}\right\rfloor$, we define the $i$-th window as the sub-sequence of $S$ that begins at link number $b_i = i\cdot \Delta$ and ends at $e_i = i\cdot\Delta + \ell/2$. Then, for a given ratio $r$, we split the $i$-th window into $X_i$ and $Y_i$ defined by $X_i = s_{b_i}, \dots, s_{r\cdot e_i}$ and $Y_i = s_{r\cdot e_i + 1}, \dots, s_{e_i}$. In other words, $X_i$ is composed of the $r\cdot \ell/2$ first links of the $i$-th window, $Y_i$ of the remaining ones.
In the following, we will use $r=0.7$ and $\Delta = \ell/100$, which implies that we always consider $50$ windows. Finally, for all $i$, we evaluate TGF with learning set $X_i$ and test set $Y_i$.

\begin{figure}[h]
\centering
\resizebox{\columnwidth}{!}{\includegraphics{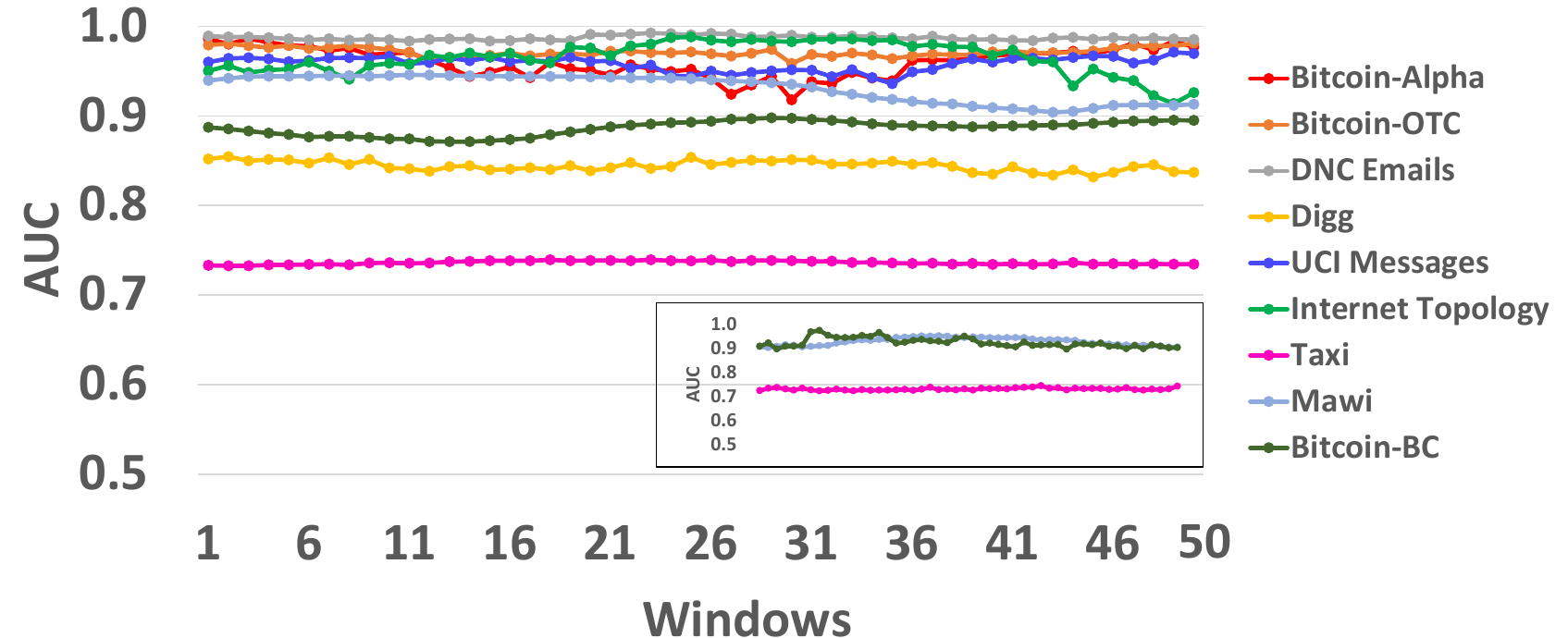}}
\caption{AUC scores for TGF with sliding windows containing $50\%$ of all links, using $H$-type history graphs of size $1000$, with $5$\% anomaly injection and learning rate $r=0.7$ in each window. The inset shows results for sliding windows containing only $1\%$ of all links in the largest datasets.}
\label{fig:streaming50}
\end{figure}

We display representative results in Figure~\ref{fig:streaming50}. In most cases, the AUC scores are strikingly stable over time. Some datasets display variations, like for instance UCI Forum for which TGF reaches better scores over time, and Mawi for which scores decline, but these variations remain very small.

In the case of huge datasets, it is interesting to consider smaller window sizes. We therefore present experiments with the same protocol as above, but with sliding windows that contain $10\%$ and $1\%$ of the links only. Obtained results are very similar in both cases. We thus display in the inset of Figure~\ref{fig:streaming50} the AUC scores obtained by TGF on our largest datasets with sliding windows that contain $1\%$ of the links only.

These experiments lead to slightly more variability than results on larger windows. However, they show that TGF performances with windows as small as $1$\% of the dataset remain close to the ones using whole datasets, and they remain quite stable.

\subsection{TGF in practice}
\label{sec:inpractice}

TGF has much flexibility: it may use various kinds of history graphs of various sizes and durations, and it may use various learning techniques with various parameters.
In practice, one is interested in a specific application area with a specific dataset. Then, one explores the choices giving the best results for this specific application, unlike the results we presented above (we used a single set of default parameters for all datasets).

We illustrate this approach on the Digg dataset, because it seems to be the most challenging case among the ones we considered: with a typical $5$\% anomaly injection, the best state-of-the-art method reaches a $0.8577$ AUC score, while TGF obtains $0.8418$ with $H_{1000}$ (Table~\ref{TableComparisonWithStateOfTheArt-Unipartite}), and $0.8988$ with a simple combination of history sizes (Figure~\ref{figCombining}). We explore here the ability of TGF to reach even better results with dedicated parameter tuning.

We first deepen the study of history size and duration influence on results, as well as the impact of used machine learning technique.

\begin{figure}[h]
\centering
\resizebox{\columnwidth}{!}{\includegraphics{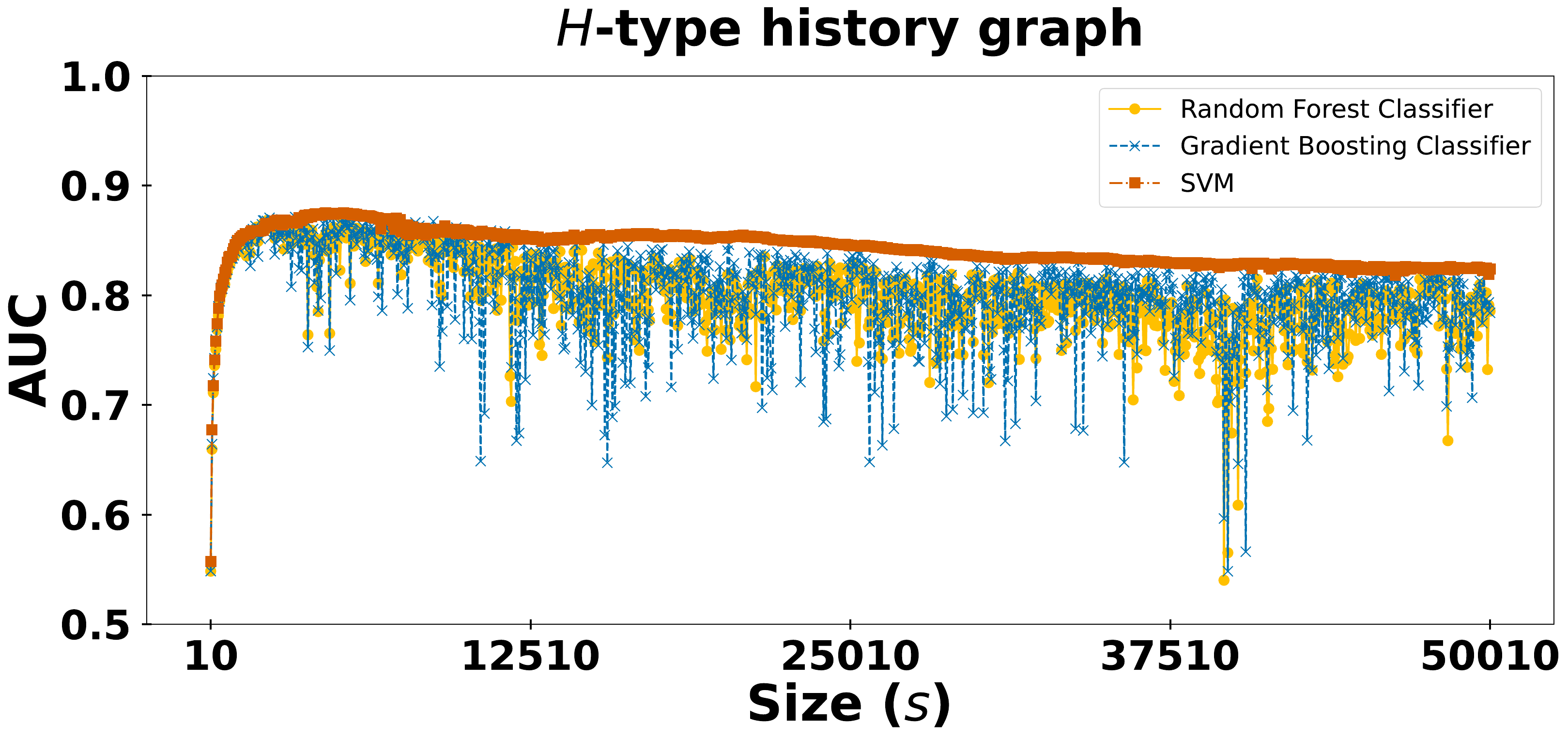}}
\smallskip  
\resizebox{\columnwidth}{!}{\includegraphics{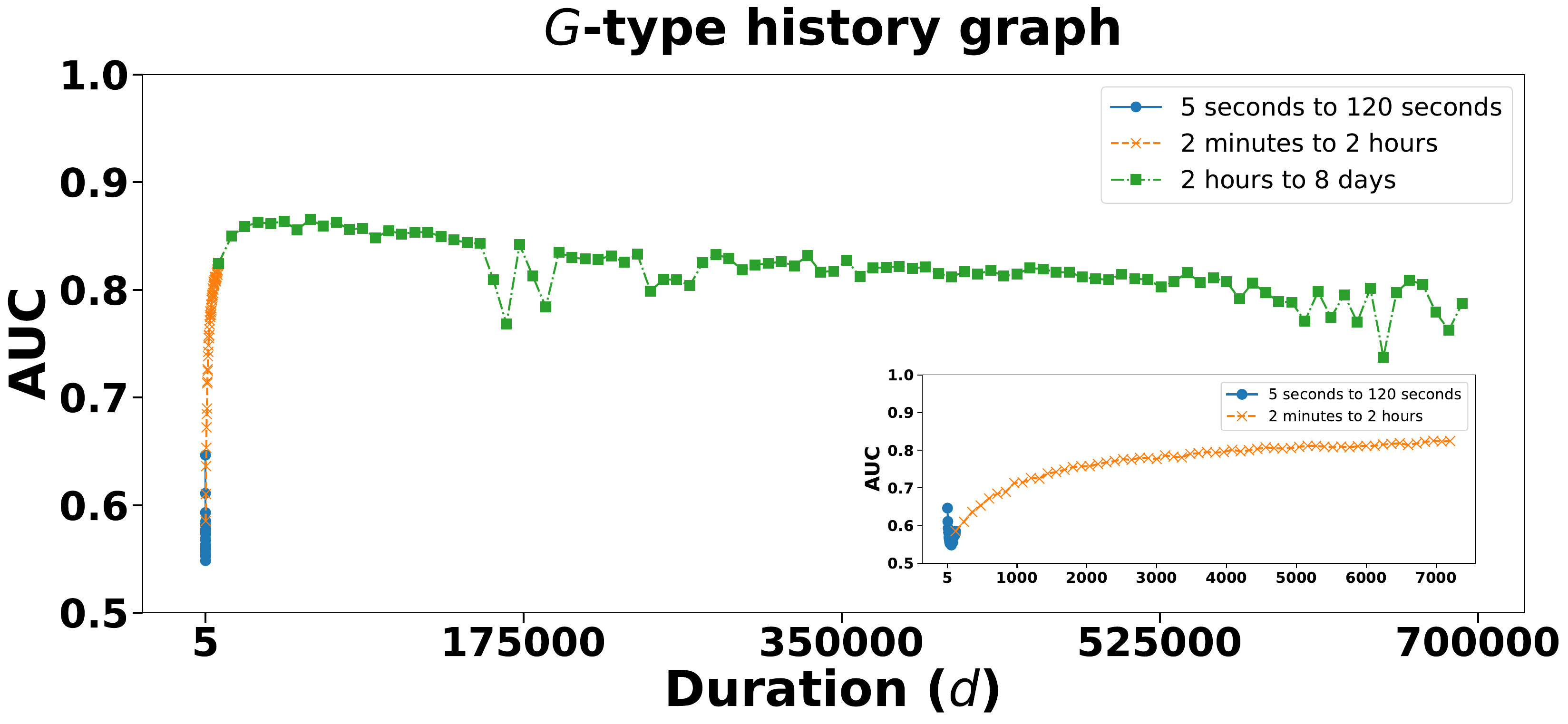}}
\caption{The impact of size $s$ and duration $d$ (horizontal axis) of the $H$-type (top) and $G$-type (bottom) history graphs on AUC scores in the Digg dataset. We also show the impact of the usage of different machine learning algorithms on $H$-type history graphs. We consider here $5$\% anomaly injection and learning with $r=0.7$.}
\label{TGFInPractice}
\end{figure}

To do so, we consider $H$-type history graphs of size 10 to 50,000 with step 10. We then perform the learning task with scikit-learn implementations of SVM and gradient boosting classifier, in addition to the random forest classifier that we used until now, each with default parameters. We display results in Figure~\ref{TGFInPractice} (top). The plot clearly shows that very small history graphs lead to poor performances, but scores very rapidly increase up to a maximum, and then slightly decrease when history size further increases. We also observe that all learning techniques perform very well. Gradient boosting classifier and random forest classifier however exhibit some variability: obtained scores drop for some specific history sizes. Interestingly, SVM gives slightly better and strikingly more stable results, but this comes with an increased learning cost.

Figure~\ref{TGFInPractice} (bottom) displays AUC scores obtained with $G$-type history graphs that span durations from 5 to 120 seconds with 5 second steps, then from $2$ minutes to $2$ hours with $2$ minute steps, and finally from $2$ hours to $8$ days with $2$ hour steps. In this experiment, we use random forest classifier as usual. It gives similar but much more stable results than with $H$-type history graphs. Therefore, using history graphs with a consistent temporal meaning is important in the Digg case.

In the experiments above, the best AUC scores obtained with $H$-type history graphs are $0.869$ with random forest learning (where $s=5160$), $0.871$ with gradient boosting classifier (where $s=3310$), and up to $0.875$ with SVM (where $s=3310$). The best AUC score obtained with $G$-type history graphs and random forest learning is $0.865$ (with $d=16$ hours). This is already an important improvement over our previous results, and significantly better than the best state-of-the-art scores, see Table~\ref{TableComparisonWithStateOfTheArt-Unipartite}.


Like in Section~\ref{sec:combination}, we now try to improve performances by combining several history graphs of the two kinds. Using all available history graphs however leads to a prohibitive number of features.
We therefore select 100 history graphs at random among the available ones of each graph type, and we use the features from these graphs only. Then, we use grid search implemented in scikit-learn in order to tune the parameters of random forest and gradient boosting classifiers. Tuning parameters of SVM is too costly.

With random forest, we obtain AUC scores of $0.8955$, $0.9007$ and $0.8829$ if we use $H$-type history graphs only, $G$-type only, or both, respectively. With gradient boosting, we obtain AUC scores of
$0.8935$, $0.8803$ and $0.9012$.

All obtained scores are excellent, above $0.88$, and we surpass the symbolic $0.9$ AUC score limit where the best state-of-the-art score was below $0.87$. In addition, using several history graphs (of any type) is always significantly better than using a single history graph. Notice however that combining both types of history graphs does not necessarily improve results, and the differences between random forest and gradient boosting performances are not significant.

\subsection{Interpretability}

TGF computes a wide variety of features, and the learning method selects the most relevant ones for anomalous link detection. The set of relevant features and their importance varies with the considered application data. As a consequence, it is useful to provide all these features to the learning method. However, understanding which features are crucial for a given application is important for interpretability purposes.

We investigate this question using a method known as feature permutation~\cite{breiman2001random}: for each feature, we compare the original detection scores to the ones obtained when the values of this feature are randomly shuffled in the test data. As a consequence these feature values do not make any sense anymore during the detection test. Performances generally decrease after permutation, and this decrease is a measure of the importance of this feature in the original score. Performing this experiment for all features leads to a ranking of features according to their importance in TGF performances, in the considered case. Here, we repeat the experiments ten times for each feature in order to gain insight on result variability.

\begin{figure}[h]
\centering
\resizebox{\columnwidth}{!}{\includegraphics{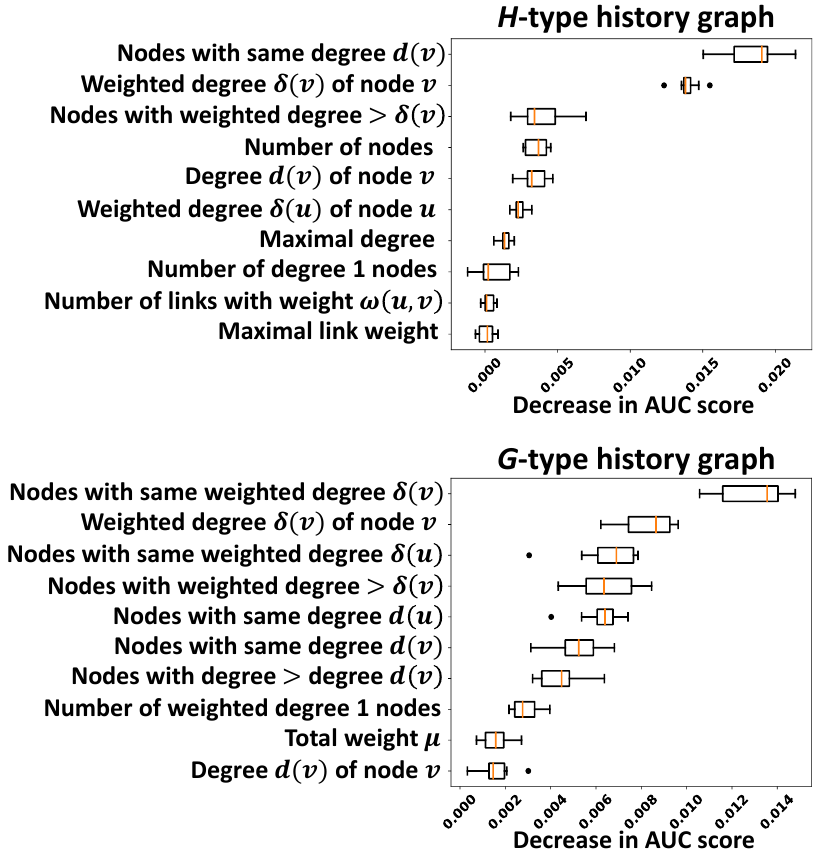}}
\caption{Feature permutation analysis for Digg with $5$\% anomaly injection, when features are computed on $H$-type history graphs with size $15\,000$ (top) and $G$-type history graphs with duration $6$ days (bottom). We display only the ten features (vertical axis) with maximal AUC score decrease (horizontal axis). In order to display box plots, we repeat the permutation ten times for each feature.}
\label{FeaturePermutationToPresent}
\end{figure}

Figure~\ref{FeaturePermutationToPresent} displays the obtained importance estimation for the ten most important features in the Digg dataset case.
It shows that not all features have the same importance: there are significant difference in decreases of AUC score induced by their permutation. Interestingly, some are important in both cases, which highlights their crucial importance for Digg. For instance, the number of nodes having the same degree as the maximal degree node in the considered link is the most important feature for the $H$-type history graph and it is the 5-th most important for the $G$-type one. Likewise, both kinds of graphs have the same second most important feature, namely the weighted degree of the maximal degree node in the link. These features already give some insight on what makes anomalous links different from normal ones, in Digg.

\begin{figure}[h]
\centering
\resizebox{\columnwidth}{!}{\includegraphics{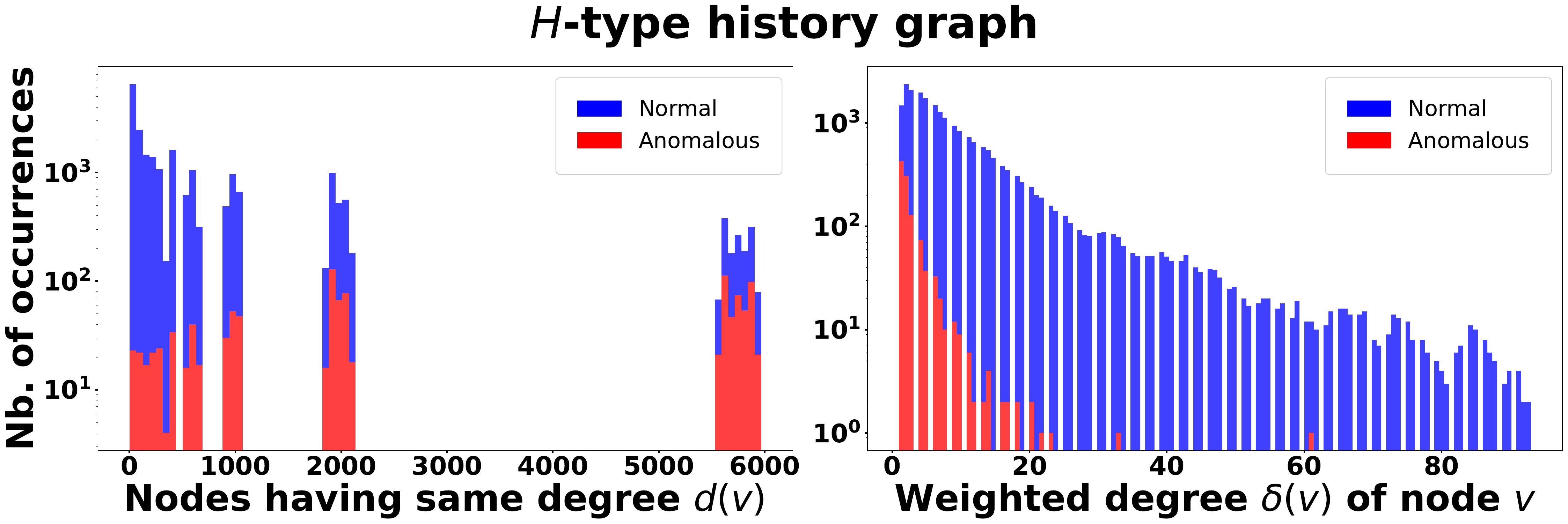}}\\
\smallskip  
\resizebox{\columnwidth}{!}{\includegraphics{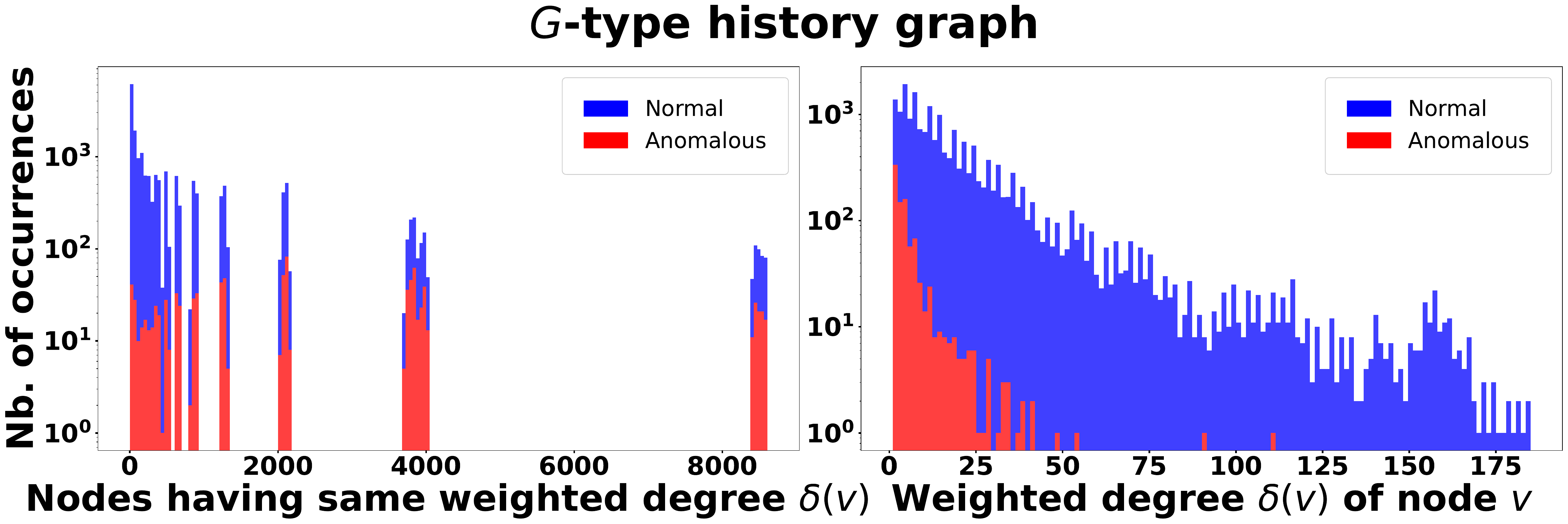}}
\caption{Distributions of values of the two most important features for the Digg dataset with 5\% injection, in $H$-type history graphs of size $15\,000$ (top) and $G$-type history graphs of duration $6$ days (bottom). The distribution for normal links is in blue, and the one for anomalous links is in red. The vertical axis is in logarithmic scale.}
\label{LogFrequency-Digg}
\end{figure}

To go further, we display in Figure~\ref{LogFrequency-Digg} the value distribution of the top two most important features for both kinds of history graphs. Our goal here is to compare these distributions for normal versus anomalous links.

First notice that distributions for normal and for anomalous links are indeed different for these features.

For instance, in $H$-type history graphs, most normal links are between nodes that do not have the same degree as most nodes (see top-left plot in Figure~\ref{LogFrequency-Digg}, notice the logarithmic vertical scale). This is due to the fact that many normal links involve high-degree nodes, which are rare in Digg. Instead, anomalous links often are between nodes that have the same degree as many nodes, since anomalous links are between randomly selected nodes, see Section~\ref{sec:injection}. Similar observations explain the differences between the number of nodes having the same weighted degree as involved nodes in $G$-type history graphs, see bottom-left plot in Figure~\ref{LogFrequency-Digg}.

The differences between normal and anomalous links are even more striking for the weighted degree of involved nodes, a crucial features for both kinds of history graphs (two rightmost plots in Figure~\ref{LogFrequency-Digg}): while many normal links are between nodes of high weighted degree, the vast majority of anomalous links are between nodes of low weighted degree. Again, this is due to the weighted degree distribution in Digg history graphs, which contains a majority of low weighted degree nodes.

These observations show how TGF results are interpretable in terms of graph features. Importantly, such simple graph features also have an interpretation in terms of what the dataset represents. In Digg, for instance, nodes are users and a link between two users means that one of them tagged a post of the other. Then, the degree of a node is the number of different users that this user tagged plus the number of users who tagged this user. Its weighted degree is the number of posts this user tagged plus the number of his or her posts that have been tagged. These quantities are classical metrics of user activity, popularity, and audience, three highly interpretable concepts for analysts. The results above show that these metrics are sufficient to very effectively make the difference between true links and randomly injected links in the Digg dataset.

\subsection{Computational costs}

TGF tackles large scale datasets with very limited space and time costs. We present here an analysis of these costs. All computations are performed on a standard personal laptop equipped with an Intel Core i7-5500 CPU @ 2.40 GHz with 8 GB RAM.

\begin{figure}[h]
\centering
\resizebox{\columnwidth}{!}{\includegraphics{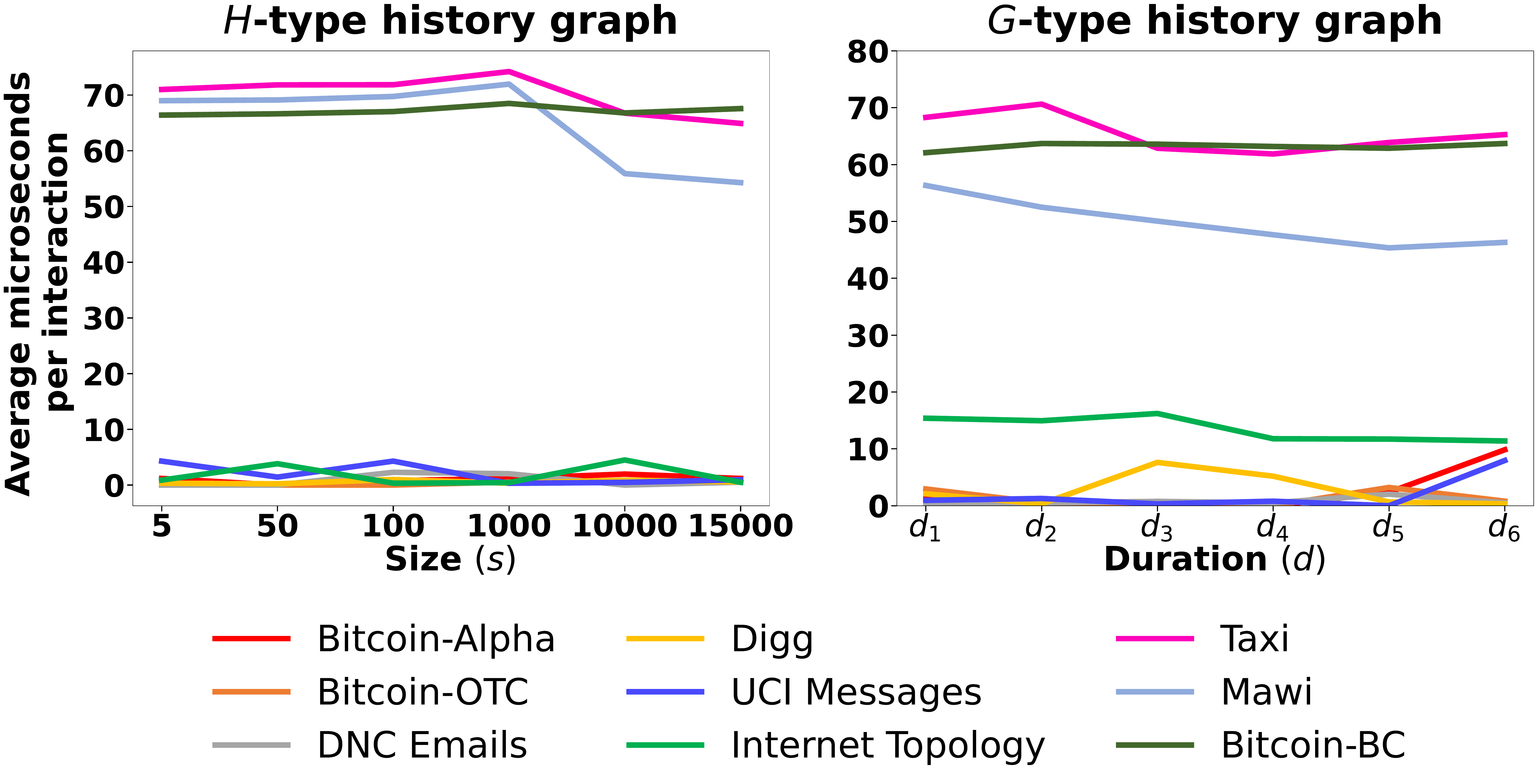}}
\caption{Average computation time (in microseconds) per link as a function of history size for $H$-type history graphs (left) and duration for $G$-type history graphs (right), for all datasets with 5\% anomaly injection.}
\label{CostsToPresent}
\end{figure}

Figure \ref{CostsToPresent} displays, for each dataset, the average computing time per transaction as a function of history type and size. This leads to several important observations.

First, in all cases, our simple python implementation processes more than 13,000 transactions per second on a typical laptop, and up to one million transactions per second, depending on the considered dataset. 
This means that typical reference datasets of hundreds of thousands links are processed almost instantly.

Second, computational costs do not grow with history size and duration. This is consistent with the fact that our graph updates and feature computations perform all their operations in $\mathcal{O}(1)$ time, as explained in Sections~\ref{sec:history-graphs} and~\ref{sec:graph-features}.

Going further, notice that the observed computation time generally is larger for largest datasets.
This is due to the fact that increasing or decreasing counters of link occurrences is much cheaper than adding or removing links in history graphs, although all these operations are in $\mathcal{O}(1)$ time. For a given history size or duration, TGF generally has to perform more link additions and removals in a large dataset than in a smaller one. Conversely it has to perform more counter updates for a small dataset than for a larger one.

For similar reasons, observed computation time often decreases when history size and duration grows: larger history sizes or durations lead to more counter updates and less link operations, and so it needs less computation time.

In addition to this analysis of feature computation costs, Table~\ref{tab:tgf_end_to_end} compares end-to-end computational costs of SLADE and TGF. This includes feature computations detailed above, as well as training and testing phases for all benchmark datasets. With TGF, the cost of these tasks reflects the performances of the chosen learning method (here, random forests with default parameters, as explained above). These experiments show that TGF consistently is very fast, and always faster than SLADE: it completes all tasks in less than one minute, whereas SLADE takes up to two hours in some cases.

\begin{table}[h!]
\centering
\caption{End-to-end computation time of SLADE and TGF $H_{1000}$ across all reference datasets, with 1\%, 5\%, and 10\% anomaly injections. Time is in hours ($h$), minutes ($m$), and seconds ($s$).}
\resizebox{.99\columnwidth}{!}{%
\begin{tabular}{l|lrrr}
\multicolumn{2}{c}{} & \multicolumn{1}{c}{1\%} & \multicolumn{1}{c}{5\%} & \multicolumn{1}{c}{10\%} \\
\hline
\multirow{2}{*}{Bitcoin-Alpha} 
& SLADE     & 0$h$ 2$m$ 21$s$ & 0$h$ 2$m$ 35$s$ & 0$h$ 2$m$ 39$s$ \\
& TGF & 0$h$ 0$m$ 5.56$s$ & 0$h$ 0$m$ 5.55$s$ & 0$h$ 0$m$ 5.53$s$ \\
\hline
\multirow{2}{*}{Bitcoin-OTC} 
& SLADE     & 0$h$ 5$m$ 30$s$ & 0$h$ 6$m$ 3$s$ & 0$h$ 6$m$ 29$s$ \\
& TGF & 0$h$ 0$m$ 11.82$s$ & 0$h$ 0$m$ 12.89$s$ & 0$h$ 0$m$ 16.71$s$ \\
\hline
\multirow{2}{*}{Digg} 
& SLADE     & 0$h$ 41$m$ 38$s$ & 0$h$ 44$m$ 55$s$ & 0$h$ 47$m$ 56$s$ \\
& TGF & 0$h$ 0$m$ 19.89$s$ & 0$h$ 0$m$ 24.76$s$ & 0$h$ 0$m$ 27.37$s$ \\
\hline
\multirow{2}{*}{DNC Emails} 
& SLADE     & 0$h$ 4$m$ 2$s$ & 0$h$ 4$m$ 28$s$ & 0$h$ 4$m$ 47$s$ \\
& TGF & 0$h$ 0$m$ 7.25$s$ & 0$h$ 0$m$ 8.14$s$ & 0$h$ 0$m$ 9.78$s$ \\
\hline
\multirow{2}{*}{Internet Topo.} 
& SLADE     & 1$h$ 52$m$ 24$s$ & 1$h$ 56$m$ 24$s$ & 2$h$ 4$m$ 20$s$ \\
& TGF & 0$h$ 0$m$ 35.31$s$ & 0$h$ 0$m$ 39.60$s$ & 0$h$ 0$m$ 42.94$s$ \\
\hline
\multirow{2}{*}{UCI Messages} 
& SLADE     & 0$h$ 7$m$ 54$s$ & 0$h$ 8$m$ 18$s$ & 0$h$ 8$m$ 49$s$ \\
& TGF & 0$h$ 0$m$ 13.37$s$ & 0$h$ 0$m$ 13.82$s$ & 0$h$ 0$m$ 16.84$s$ \\
\end{tabular}%
}
\label{tab:tgf_end_to_end}
\end{table}

\section{Conclusion} \label{sec:conclusion}

Trivial graph features and classical learning are enough for detecting anomalies, as long as they are similar to random links. As a consequence, it makes no sense to continue the line of works that develop overly complex detection methods tailored for such anomalies. Instead, detection methods should now target more complex kinds of anomalies, real or injected with advanced models. This calls for better characterizations of real anomalies, which raises again the issue of limited labeled data availability. Another promising option is to develop anomaly models more subtle than purely random ones. In these directions, most remains to be done.

TGF has a great potential for extensions in order to detect these more complex anomalies. Indeed, it readily handles various kinds of graphs (the bipartite case is detailed in online appendix \cite{TGFweb}) as well as domain-specific data (one only has to append their features to ours before learning). In addition, many less trivial but still very efficient graph features may easily be added, like for instance features leveraging local density or distances.

\section*{Acknowledgment}

This work is funded in part by CNRS through the MITI interdisciplinary programs and by ANR (French National Agency of Research) under the FiT LabCom grant.


\bibliographystyle{IEEEtran}
\bibliography{IEEEabrv,sample}

\end{document}

%% file: history.tikz
\begin{tikzpicture}

\newcommand\timecolor{NavyBlue}
\newcounter{i}


\setcounter{i}{0} 
\foreach \X in {d,c,b,a}{
 \addtocounter{i}{1}
 \coordinate (\X) at (-.5,\thei);
 \draw[font=\large] (\X) node {$\X$};
 \foreach \Y in {0,1,2,3,4,5,6,7,8,9,10}{
  \coordinate (\Y\X) at (\Y,\thei);
  }
 }

\foreach \X in {a,b,c,d}{ 
 \foreach \Y in {0,1,2,3,4,5,6,7,8,9,10}{
  \draw [color=gray!90] (\Y\X) circle (.05);
  }
 }

\tikzset{ls-node/.style={color=BurntOrange,fill=BurntOrange!50,opacity=1,thick}};
\tikzset{ls-link/.style={color=BurntOrange,opacity=1,very thick}};
\foreach \X in {0b,0d,1c,4b,6c,7c,8d,9d,9d,9b}{ 
 \draw [ls-link] (\X) -- ++(0,1);
 \draw [ls-node] (\X) circle (.08);
 \draw [ls-node] ($ (\X) + (0,1) $) circle (.08);
}
\draw [ls-link,bend right] (3d) to (3b);
\draw [ls-node] (3b) circle (.08);
\draw [ls-node] (3d) circle (.08);
\draw [ls-link,color=Red,thin] (10c) -- ++(0,1);
\draw [ls-node,color=Red,fill=White] (10c) circle (.08);
\draw [ls-node,color=Red,fill=White] ($ (10c) + (0,1) $) circle (.08);

\draw [black,thin,->] (0,0.5) -- (10.5,0.5);
\draw (10.5,0.5) node[right] {$t$};
\foreach \X in {0,1,2,3,4,5,6,7,8,9,10}{
 \draw (\X,0.5) -- (\X,0.4);
 \draw (\X,0.2) node[\timecolor] {$\X$};
}


\tikzstyle{g-node}=[circle,inner sep=3,draw,very thin,white,fill=white,text=Black,minimum width,minimum height,font=\large]
\tikzstyle{g-link}=[very thick,ForestGreen]
\tikzstyle{g-link-label}=[midway]

\begin{scope}[shift={(-.5,-4.5)}]
\node[g-node] (Ga) at (1,4) {$a$};
\node[g-node] (Gb) at (1,2.5) {$b$};
\node[g-node] (Gc) at (0.1,1) {$c$};
\node[g-node] (Gd) at (1.9,1) {$d$};
\draw[g-link] (Ga)--(Gb) node[g-link-label,right]{$1$};
\draw[g-link] (Gb)--(Gc) node[g-link-label,pos=.4,left]{$1$};
\draw[g-link] (Gc)--(Gd) node[g-link-label,below]{$2$};
\draw (1,0) node[font=\Large] {$G_3$};
\end{scope}

\begin{scope}[shift={(2.5,-4.5)}]
\node[g-node] (Ga) at (1,4) {$a$};
\node[g-node] (Gb) at (1,2.5) {$b$};
\node[g-node] (Gc) at (0.1,1) {$c$};
\node[g-node] (Gd) at (1.9,1) {$d$};
\draw[g-link] (Ga)--(Gb) node[g-link-label,right]{$2$};
\draw[g-link] (Gb)--(Gc) node[g-link-label,pos=.4,left]{$2$};
\draw[g-link] (Gb)--(Gd) node[g-link-label,pos=.4,right]{$1$};
\draw[g-link] (Gc)--(Gd) node[g-link-label,below]{$2$};
\draw (1,0) node[font=\Large] {$G_8$};
\end{scope}

\begin{scope}[shift={(5.5,-4.5)}]
\node[g-node] (Ga) at (1,4) {$a$};
\node[g-node] (Gb) at (1,2.5) {$b$};
\node[g-node] (Gc) at (0.1,1) {$c$};
\node[g-node] (Gd) at (1.9,1) {$d$};
\draw[g-link] (Ga)--(Gb) node[g-link-label,right]{$1$};
\draw[g-link] (Gc)--(Gd) node[g-link-label,below]{$2$};
\draw (1,0) node[font=\Large] {$H_3$};
\end{scope}

\begin{scope}[shift={(8.5,-4.5)}]
\node[g-node] (Ga) at (1,4) {$a$};
\node[g-node] (Gb) at (1,2.5) {$b$};
\node[g-node] (Gc) at (0.1,1) {$c$};
\node[g-node] (Gd) at (1.9,1) {$d$};
\draw[g-link] (Ga)--(Gb) node[g-link-label,right]{$2$};
\draw[g-link] (Gb)--(Gc) node[g-link-label,pos=.4,left]{$3$};
\draw[g-link] (Gb)--(Gd) node[g-link-label,pos=.4,right]{$1$};
\draw[g-link] (Gc)--(Gd) node[g-link-label,below]{$2$};
\draw (1,0) node[font=\Large] {$H_8$};
\end{scope}

\end{tikzpicture}